# Online Alternating Direction Method

**Huahua Wang** HUWANG@CS.UMN.EDU
**Arindam Banerjee** BANERJEE@CS.UMN.EDU
Dept of Computer Science & Engg, University of Minnesota, Twin Cities

## Abstract

Online optimization has emerged as powerful tool in large scale optimization. In this paper, we introduce efficient online algorithms based on the alternating directions method (ADM). We introduce a new proof technique for ADM in the batch setting, which yields the $O(1/T)$ convergence rate of ADM and forms the basis of regret analysis in the online setting. We consider two scenarios in the online setting, based on whether the solution needs to lie in the feasible set or not. In both settings, we establish regret bounds for both the objective function as well as constraint violation for general and strongly convex functions. Preliminary results are presented to illustrate the performance of the proposed algorithms.

## 1. Introduction

In recent years, online learning (Zinkevich, 2003; Hazan et al., 2007) and its batch counterpart stochastic gradient descent (Juditsky et al., 2009) has contributed substantially to advances in large scale optimization techniques for machine learning. Online convex optimization has been generalized to handle time-varying and non-smooth convex functions (Duchi et al., 2010; Duchi & Singer, 2009; Xiao, 2010). Distributed optimization, where the problem is divided into parts on which progress can be made in parallel, has also contributed to advances in large scale optimization (Boyd et al., 2010; Bertsekas & Tsitsiklis, 1989; Censor & Zenios, 1998).

Important advances have been made based on the above ideas in the recent literature. Composite objective mirror descent (COMID) (Duchi et al., 2010) generalizes mirror descent (Beck & Teboulle, 2003) to the online setting. COMID also includes certain other proximal splitting methods such as FOBOS (Duchi & Singer, 2009) as special cases. Regularized dual averaging (RDA) (Xiao, 2010) generalizes dual averaging (Nesterov, 2009) to online and composite optimization, and can be used for distributed optimization (Duchi et al., 2011).

First introduced in (Gabay & Mercier, 1976), the alternating direction method (ADM) has become popular in recent years due to its ease of applicability and empirical performance in a wide variety of problems, including composite objectives (Boyd et al., 2010; Eckstein & Bertsekas, 1992; Lin et al., 2009). The proof of convergence of ADM can be found in (Eckstein & Bertsekas, 1992; Boyd et al., 2010), although the rate of convergence rate has not been established. For further understanding of ADM, we refer the readers to the comprehensive review by (Boyd et al., 2010). An advantage of ADM is that it can handle linear equality constraints of the form $\{\mathbf{x}, \mathbf{z} | \mathbf{A}\mathbf{x} + \mathbf{B}\mathbf{z} = \mathbf{c}\}$, which makes distributed optimization by variable splitting in a batch setting straightforward (Boyd et al., 2010). However, in an online or stochastic gradient descent setting, one obtains a double-loop algorithm where the inner loop ADM iterations have to be run till convergence after every new data point or function is revealed. As a result, ADM has not yet been generalized to the online setting.

In this paper, we consider optimization problems of the following form:

$$\min_{\mathbf{x}\in\mathcal{X},\mathbf{z}\in\mathcal{Z}} \sum_{t=1}^{T} (f_t(\mathbf{x}) + g(\mathbf{z})) \quad \text{s.t.} \quad \mathbf{A}\mathbf{x} + \mathbf{B}\mathbf{z} = \mathbf{c}\ , \quad (1)$$

where the functions $f_t, g$ are (non-smooth) convex functions, $\mathbf{A} \in \mathbb{R}^{m\times n_1}, \mathbf{B} \in \mathbb{R}^{m\times n_2}, \mathbf{c} \in \mathbb{R}^m$, $\mathbf{x} \in \mathcal{X} \in \mathbb{R}^{n_1\times 1}, \mathbf{z} \in \mathcal{Z} \in \mathbb{R}^{n_2\times 1}$, where $\mathcal{X}$ and $\mathcal{Z}$ are convex sets. In the sequel, we drop the convex sets $\mathcal{X}$ and $\mathcal{Z}$ for ease of exposition, noting that one can consider $g$ and other additive functions to be the indicators of suitable convex feasible sets. The problem is studied both in the batch setting, where $f_t = f$, and in the online setting for time-varying $f_t$. We introduce a new proof technique for ADM in the batch setting, which establishes a $O(1/T)$ convergence rate of ADM based on variational inequalities (Facchinei & Pang, 2003). Further, the convergence analysis for the batch setting forms the basis of regret analysis in the online setting. We consider two scenarios in the online setting, based on whether or not the solution needs to lie in the feasible set in every iteration.

We propose efficient online ADM (OADM) algorithms for both scenarios which make a single pass through the update equations and avoid a double loop algorithm. In the online setting, while a single pass through the ADM update equations is not guaranteed to satisfy the linear constraints $\mathbf{A}\mathbf{x}_t + \mathbf{B}\mathbf{z}_t = \mathbf{c}$, we consider two types of regret: regret in the *objective* as well as regret in *constraint violation*. We establish both types of regret bounds for general and strongly convex functions. We also present preliminary experimental results illustrating the performance of the proposed OADM algorithms in comparison with FOBOS and RDA (Duchi & Singer, 2009; Xiao, 2010).

The key advantage of the OADM algorithms can be summarized as follows: Like COMID and RDA, OADM can solve online composite optimization problems, matching the regret bounds for existing methods. The ability to additionally handle linear equality constraints of the form $\mathbf{A}\mathbf{x} + \mathbf{B}\mathbf{z} = \mathbf{c}$ makes non-trivial variable splitting possible yielding efficient distributed online optimization algorithms based on OADM. Further, the notion of regret in both the objective as well as constraints may contribute towards development of suitable analysis tools for online constrained optimization problems (Mannor & Tsitsiklis, 2006; Mahdavi et al., 2011).

The rest of the paper is organized as follows. In Section 2, we analyze batch ADM and establish its convergence rate. In Section 3, we introduce the online optimization problem with linear constraints. The OADM algorithm is also given in Section 3. In Sections 4 and 5, we present the regret analysis in two different scenarios based on the constraints. We discuss connections to related work in Section 6, present preliminary experimental results in Section 7, and conclude in Section 8.

## 2. Analysis for Batch ADM

We consider the batch ADM problem (1) where $f_t$ is fixed. The augmented Lagrangian for (1) is

$$L_\rho(\mathbf{x},\mathbf{y},\mathbf{z}) = f(\mathbf{x}) + g(\mathbf{z}) + \langle \mathbf{y}, \mathbf{A}\mathbf{x} + \mathbf{B}\mathbf{z} - \mathbf{c}\rangle + \frac{\rho}{2}\|\mathbf{A}\mathbf{x} + \mathbf{B}\mathbf{z} - \mathbf{c}\|^2, \quad (2)$$

where $\mathbf{z}$ is the primal variable and $\mathbf{y}$ is the dual variable, $\rho > 0$ is the penalty parameter. Batch ADM executes the following three steps iteratively till convergence (Boyd et al., 2010):

$$\mathbf{x}_{t+1} = \operatorname*{argmin}_{\mathbf{x}} f(\mathbf{x}) + \langle \mathbf{y}_t, \mathbf{A}\mathbf{x} + \mathbf{B}\mathbf{z}_t - \mathbf{c}\rangle + \frac{\rho}{2}\|\mathbf{A}\mathbf{x} + \mathbf{B}\mathbf{z}_t - \mathbf{c}\|^2, \quad (3)$$

$$\mathbf{z}_{t+1} = \operatorname*{argmin}_{\mathbf{z}} g(\mathbf{z}) + \langle \mathbf{y}_t, \mathbf{A}\mathbf{x}_{t+1} + \mathbf{B}\mathbf{z} - \mathbf{c}\rangle + \frac{\rho}{2}\|\mathbf{A}\mathbf{x}_{t+1} + \mathbf{B}\mathbf{z} - \mathbf{c}\|^2, \quad (4)$$

$$\mathbf{y}_{t+1} = \mathbf{y}_t + \rho(\mathbf{A}\mathbf{x}_{t+1} + \mathbf{B}\mathbf{z}_{t+1} - \mathbf{c}) . \quad (5)$$

At step $(t+1)$, the equality constraint is not necessarily satisfied in ADM. However, one can show that the equality constraint is satisfied in the long run such that $\lim_{t\to\infty} \mathbf{A}\mathbf{x}_t + \mathbf{B}\mathbf{z}_t - \mathbf{c} \to \mathbf{0}$. We first analyze the convergence of objective and constraint separately using a new proof technique, which plays an important role for the regret analysis in the online setting. Then, a joint analysis of the objective and constraint using a variational inequality (Facchinei & Pang, 2003) establishes the $O(1/T)$ convergence rate for ADM.

Without loss of generality, we assume that $\mathbf{z}_0 = 0, \mathbf{y}_0 = 0$. Denote $\|\mathbf{y}^*\|_2 = D_{\mathbf{y}}, \|\mathbf{z}^*\|_2 = D_{\mathbf{z}}$ and $\lambda_{\max}^{\mathbf{B}}$ as the largest eigenvalue of $\mathbf{B}^T\mathbf{B}$.

### 2.1. Bounds for Objective and Constraints

The following theorem shows that both the cumulative objective difference w.r.t. the optimal and the cumulative norms of the constraints, known as the primal and dual residuals (Boyd et al., 2010), are bounded by constants independent of the number of iterations $T$.

**Theorem 1** *Let the sequences $\{\mathbf{x}_t, \mathbf{z}_t, \mathbf{y}_t\}$ be generated by ADM. For any $\mathbf{x}^*, \mathbf{z}^*$ satisfying $\mathbf{A}\mathbf{x}^* + \mathbf{B}\mathbf{z}^* = \mathbf{c}$, for any $T$, we have*

$$\sum_{t=0}^{T} [f(\mathbf{x}_{t+1}) + g(\mathbf{z}_{t+1}) - (f(\mathbf{x}^*) + g(\mathbf{z}^*))] \leq \frac{\lambda_{\max}^{\mathbf{B}} D_{\mathbf{z}}^2 \rho}{2} , \quad (6)$$

$$\sum_{t=0}^{T} \|\mathbf{A}\mathbf{x}_{t+1} + \mathbf{B}\mathbf{z}_{t+1} - \mathbf{c}\|_2^2 + \|\mathbf{B}\mathbf{z}_{t+1} - \mathbf{B}\mathbf{z}_t\|_2^2 \leq \lambda_{\max}^{\mathbf{B}} D_{\mathbf{z}}^2 + \frac{D_{\mathbf{y}}^2}{\rho^2} . \quad (7)$$

It is easy to verify that the KKT conditions of the augmented lagrangian (2) hold if (7) holds. The convergence of equality constraint and primal residual implies the convergence of ADM. A result similar to (7) has been shown in (Boyd et al., 2010), but our proof is different and self-contained along with (6). Although (6) shows that the objective value converges to the optimal value, $\mathbf{x}_{t+1}, \mathbf{z}_{t+1}$ need not be feasible and the equality constraint is not necessarily satisfied.

### 2.2. Rate of Convergence of ADM

We now prove the $O(1/T)$ convergence rate for ADM using a variational inequality (VI) based on the Lagrangian given in (2). Let $\Omega = \mathcal{X} \times \mathcal{Z} \times \mathbb{R}^m$. Any $\mathbf{w}^* = (\mathbf{x}^*, \mathbf{z}^*, \mathbf{y}^*) \in \Omega$ solves the original problem in (1) optimally if it satisfies the following variational inequality (Facchinei & Pang, 2003; Nemirovski, 2004):

$$\forall \mathbf{w} \in \Omega , \quad h(\mathbf{w}) - h(\mathbf{w}^*) + (\mathbf{w} - \mathbf{w}^*)^T F(\mathbf{w}^*) \geq 0 , \quad (8)$$

where $F(\mathbf{w})^T = [\mathbf{y}^T\mathbf{A} \quad \mathbf{y}^T\mathbf{B} \quad -(\mathbf{A}\mathbf{x} + \mathbf{B}\mathbf{z} - \mathbf{c})^T]$ is the gradient of the last term of the Lagrangian, and $h(\mathbf{w}) = f(\mathbf{x}) + g(\mathbf{z})$. Then, $\tilde{\mathbf{w}} = (\tilde{\mathbf{x}}, \tilde{\mathbf{z}}, \tilde{\mathbf{y}})$ approximately solves the problem with accuracy $\epsilon$ if it satisfies

$$\forall \mathbf{w} \in \Omega\,, \quad h(\tilde{\mathbf{w}}) - h(\mathbf{w}) + (\tilde{\mathbf{w}} - \mathbf{w})^T F(\tilde{\mathbf{w}}) \leq \epsilon\,. \quad (9)$$

We show that after $T$ iterations, the average $\bar{\mathbf{w}}_T = \frac{1}{T}\sum_{t=1}^T \mathbf{w}_t$, where $\mathbf{w}_t = (\mathbf{x}_t, \mathbf{z}_t, \mathbf{y}_t)$ are from (3)-(5), satisfies the above inequality with $\epsilon = O(1/T)$.

**Theorem 2** *Let* $\bar{\mathbf{w}}_T = \frac{1}{T}\sum_{t=1}^T \mathbf{w}_t$, *where* $\mathbf{w}_t = (\mathbf{x}_t, \mathbf{z}_t, \mathbf{y}_t)$ *from (3)-(5). Then,*

$$\forall \mathbf{w} \in \Omega, \;\; h(\bar{\mathbf{w}}_T) - h(\mathbf{w}) + (\bar{\mathbf{w}}_T - \mathbf{w})^T F(\bar{\mathbf{w}}_T) \leq O\left(\frac{1}{T}\right).$$

## 3. Online ADM

In this section, we extend the ADM to the online learning setting. Specifically, we focus on using online ADM (OADM) to solve the problem in (1). For our analysis, $\mathbf{A}$ and $\mathbf{B}$ are assumed to be fixed. At round $t$, we consider solving the following regularized optimization problem:

$$\mathbf{x}_{t+1} = \operatorname*{argmin}_{\mathbf{Ax}+\mathbf{Bz}=\mathbf{c}} f_t(\mathbf{x}) + g(\mathbf{z}) + \eta B_\phi(\mathbf{x}, \mathbf{x}_t)\,, \quad (10)$$

where $\eta \geq 0$ is a learning rate and Bregman divergence $B_\phi(\mathbf{x}, \mathbf{x}_t) \geq \frac{\alpha}{2}\|\mathbf{x} - \mathbf{x}_t\|_2^2$. If the above problem is solved in every step, standard analysis techniques (Hazan et al., 2007) can be suitably adopted to obtain sublinear regret bounds. While (10) can be solved by batch ADM, we essentially obtain a double loop algorithm where the function $f_t$ changes in the outer loop and the inner loop runs ADM iteratively till convergence so that the constraints are satisfied. Note that existing online methods, such as projected gradient descent and variants (Hazan et al., 2007; Duchi et al., 2010) do assume a black-box approach for projecting onto the feasible set, which for linear constraints may require iterative cyclic projections (Censor & Zenios, 1998).

For our analysis, instead of requiring the equality constraints to be satisfied at each time $t$, we only require the equality constraints to be satisfied in the long run, with a notion of regret associated with constraints. In particular, we consider the following online learning problem:

$$\min_{\mathbf{x}_t, \mathbf{z}_t} \sum_{t=0}^T f_t(\mathbf{x}_t) + g(\mathbf{z}_t) - \min_{\mathbf{Ax}+\mathbf{Bz}=\mathbf{c}} \sum_{t=0}^T f_t(\mathbf{x}) + g(\mathbf{z})$$

$$s.t. \quad \sum_{t=1}^T \|\mathbf{A}\mathbf{x}_t + \mathbf{B}\mathbf{z}_t - \mathbf{c}\|_2^2 = o(T)\,, \quad (11)$$

so that the cumulative constraint violation is sublinear in $T$. The augmented lagrangian function of (10) at time $t$ is

$$L_t(\mathbf{x}, \mathbf{y}, \mathbf{z}) = f_t(\mathbf{x}) + g(\mathbf{z}) + \langle \mathbf{y}, \mathbf{Ax} + \mathbf{Bz} - \mathbf{c}\rangle + \eta B_\phi(\mathbf{x}, \mathbf{x}_t) + \frac{\rho}{2}\|\mathbf{Ax} + \mathbf{Bz} - \mathbf{c}\|^2\,. \quad (12)$$

At time $t$, our algorithm consists of just one pass through the following three update steps:

$$\mathbf{x}_{t+1} = \operatorname*{argmin}_{\mathbf{x}} f_t(\mathbf{x}) + \langle \mathbf{y}_t, \mathbf{Ax} + \mathbf{B}\mathbf{z}_t - \mathbf{c}\rangle + \frac{\rho}{2}\|\mathbf{Ax} + \mathbf{B}\mathbf{z}_t - \mathbf{c}\|^2 + \eta B_\phi(\mathbf{x}, \mathbf{x}_t)\,, \quad (13)$$

$$\mathbf{z}_{t+1} = \operatorname*{argmin}_{\mathbf{z}} g(\mathbf{z}) + \langle \mathbf{y}_t, \mathbf{A}\mathbf{x}_{t+1} + \mathbf{Bz} - \mathbf{c}\rangle + \frac{\rho}{2}\|\mathbf{A}\mathbf{x}_{t+1} + \mathbf{Bz} - \mathbf{c}\|^2\,, \quad (14)$$

$$\mathbf{y}_{t+1} = \mathbf{y}_t + \rho(\mathbf{A}\mathbf{x}_{t+1} + \mathbf{B}\mathbf{z}_{t+1} - \mathbf{c})\,. \quad (15)$$

The $\mathbf{x}$-update (13) has two penalty terms: a quadratic term and a Bregman divergence. If the Bregman divergence is not a quadratic function, it may be difficult to solve $\mathbf{x}$ efficiently. A common way is to linearize the objective such that

$$\mathbf{x}_{t+1} = \operatorname*{argmin}_{\mathbf{x}} \langle f_t'(\mathbf{x}_t) + \mathbf{A}^T\{\mathbf{y}_t + \rho(\mathbf{A}\mathbf{x}_t + \mathbf{B}\mathbf{z}_t - \mathbf{c})\}, \mathbf{x} - \mathbf{x}_t\rangle + \eta B_\phi(\mathbf{x}, \mathbf{x}_t)\,. \quad (16)$$

(16) is known as inexact ADM (Boyd et al., 2010) if $\phi$ is a quadratic function. In the sequel, we focus on the algorithm using (13).

Operationally, in round $t$, the algorithm presents a solution $\{\mathbf{x}_t, \mathbf{z}_t\}$ as well as $\mathbf{y}_t$. Then, nature reveals function $f_t$ and we encounter two types of losses. The first type is the traditional loss measured by $f_t(\mathbf{x}_t) + g(\mathbf{z}_t)$. The second type is the residual of constraint violation, i.e., $\|\mathbf{A}\mathbf{x}_t + \mathbf{B}\mathbf{z}_t - \mathbf{c}\|^2$. The goal is to establish sublinear regret bounds for both the objective and the constraint violation, which we do in Section 4. We consider another scenario, where in round $t$, we use a solution $\{\hat{\mathbf{x}}_t, \mathbf{z}_t\}$ based on $\mathbf{z}_t$ such that $\mathbf{A}\hat{\mathbf{x}} + \mathbf{B}\mathbf{z}_t = \mathbf{c}$. While $(\hat{\mathbf{x}}_t, \mathbf{z}_t)$ satisfies the constraint by design, the goal is to establish sublinear regret of the objective $f_t(\hat{\mathbf{x}}_t) + g(\mathbf{z}_t)$ as well as the constraint violation for the true $(\mathbf{x}_t, \mathbf{z}_t)$. For the second scenario, we use $\eta = 0$ in (13) and present the results in Section 5. As the updates include the primal and dual variables, in line with batch ADM, we use a stronger regret $R^c(T) = \sum_{t=1}^T R_t^c$ for constraint violation based on both primal and dual residuals, where

$$R_t^c = \|\mathbf{A}\mathbf{x}_{t+1} + \mathbf{B}\mathbf{z}_{t+1} - \mathbf{c}\|_2^2 + \|\mathbf{B}\mathbf{z}_{t+1} - \mathbf{B}\mathbf{z}_t\|_2^2\,. \quad (17)$$

Before getting into the regret analysis, we discuss some example problems which can be solved using OADM. Like FOBOS and RDA, OADM can deal with machine learning methods where $f_t$ is a loss function and $g$ is a regularizer, e.g. $\ell_1$ or mixed norm, or an indicator function of a convex set. Examples include generalized lasso and group lasso (Boyd et al., 2010; Tibshirani, 1996; Xiao, 2010). OADM can also solve linear programs, e.g. MAP LP relaxation (Meshi & Globerson, 2011) and LP decoding (Barman et al., 2012), and non-smooth optimization, e.g. robust PCA (Lin et al., 2009) where $f_t$ is nuclear norm and $g$ is $\ell_1$ norm. Another promising scenario for OADM is consensus optimization (Boyd et al., 2010) where distributed local variables are updated separately and reach a global consensus in the long run. More examples can be found in (Boyd et al., 2010).

In the sequel, we need the following assumptions:

(1) The norm of subgradient of $f_t(\mathbf{x})$ is bounded by $G_f$.

(2) We assume $g(\mathbf{z}_0) = 0$ and $g(\mathbf{z}) \geq 0$.

(3) $\mathbf{x}_0 = \mathbf{0}, \mathbf{y}_0 = \mathbf{0}, \mathbf{z}_0 = \mathbf{0}$. For any $\mathbf{x}^*, \mathbf{z}^*$ satisfying $\mathbf{A}\mathbf{x}^* + \mathbf{B}\mathbf{z}^* = \mathbf{c}$, $B_\phi(\mathbf{x}^*, \mathbf{0}) = D_{\mathbf{x}}^2$, $\|\mathbf{z}^*\|_2 = D_{\mathbf{z}}$.

(4) For any $t$, $f_t(\mathbf{x}_{t+1}) + g(\mathbf{z}_{t+1}) - (f_t(\mathbf{z}^*) + g(\mathbf{z}^*)) \geq -F$, which is true if the functions are lower bounded or Lipschitz continuous in the convex set (Mahdavi et al., 2011).

# 4. Regret Analysis of OADM

As discussed in Section 3, we consider two types of regret in OADM. The first type is the regret of the objective based on variable splitting, i.e.,

$$R_1(T) = \sum_{t=0}^{T} f_t(\mathbf{x}_t) + g(\mathbf{z}_t) - \min_{\mathbf{A}\mathbf{x}+\mathbf{B}\mathbf{z}=\mathbf{c}} \sum_{t=0}^{T} f_t(\mathbf{x}) + g(\mathbf{z}) \,. \quad (18)$$

Aside from using splitting variables, $R_1$ is the standard regret in the online learning setting. The second is the regret of the constraint violation $R^c$ defined in (17).

## 4.1. General Convex Functions

The following establishes the regret bounds for OADM.

**Theorem 3** *Let the sequences $\{\mathbf{x}_t, \mathbf{z}_t, \mathbf{y}_t\}$ be generated by OADM and assumptions (1)-(4) hold. For any $\mathbf{x}^*, \mathbf{z}^*$ satisfying $\mathbf{A}\mathbf{x}^* + \mathbf{B}\mathbf{z}^* = \mathbf{c}$, setting $\eta = \frac{G_f\sqrt{T}}{D_{\mathbf{x}}\sqrt{2\alpha}}$ and $\rho = \sqrt{T}$, we have*

$$R_1(T) \leq \lambda_{\max}^{\mathbf{B}} D_{\mathbf{z}}^2 \sqrt{T}/2 + \sqrt{2} G_f D_{\mathbf{x}} \sqrt{T}/\sqrt{\alpha} \,,$$
$$R^c(T) \leq \lambda_{\max}^{\mathbf{B}} D_{\mathbf{z}}^2 + \sqrt{2} D_{\mathbf{x}} G_f/\sqrt{\alpha} + 2F\sqrt{T} \,.$$

Note the bounds are achieved without any explicit assumptions on $\mathbf{A}, \mathbf{B}, \mathbf{c}$.[1] The subgradient of $f_t$ is required to be bounded, but the subgradient of $g$ is not necessarily bounded. Thus, the bounds hold for the case that $g$ is an indicator function of a convex set. In addition to the $O(\sqrt{T})$ regret bound, OADM achieves the $O(\sqrt{T})$ bound for the constraint violation, which is not existent in the start-of-the-art online learning algorithms (Duchi et al., 2010; Duchi & Singer, 2009; Xiao, 2010), since they do not explicitly handle linear constraints of the form $\mathbf{A}\mathbf{x}_t + \mathbf{B}\mathbf{z} = \mathbf{c}$.

[1] We do assume that $\mathbf{A}\mathbf{x} + \mathbf{B}\mathbf{z} = \mathbf{c}$ is feasible.

The bound for $R^c$ could be reduced to a constant if additional assumptions on $\mathbf{B}$ and the subgradient of $g$ are satisfied.

## 4.2. Strongly Convex Functions

We assume both $f_t(\mathbf{x})$ and $g$ are strongly convex. Specifically, we assume $f_t(\mathbf{x})$ is $\beta_1$-strongly convex with respect to a differentiable function $\phi$, i.e.,

$$f_t(\mathbf{x}^*) \geq f_t(\mathbf{x}) + \langle f_t'(\mathbf{x}), \mathbf{x}^* - \mathbf{x}\rangle + \beta_1 B_\phi(\mathbf{x}^*, \mathbf{x}_{t+1}) \,, \quad (19)$$

where $\beta_1 > 0$, and $g$ is a $\beta_2$-strongly convex function, i.e.,

$$g(\mathbf{z}^*) \geq g(\mathbf{z}) + \langle g'(\mathbf{z}), \mathbf{z}^* - \mathbf{z}\rangle + \frac{\beta_2}{2}\|\mathbf{z}^* - \mathbf{z}_{t+1}\|_2^2 \,, \quad (20)$$

where $\beta_2 > 0$. Then, logarithmic regret bounds can be established.

**Theorem 4** *Let assumptions (1)-(4) hold. Assume $f_t(\mathbf{x})$ and $g$ are strongly convex given in (19) and (20). For any $\mathbf{x}^*, \mathbf{z}^*$ satisfying $\mathbf{A}\mathbf{x}^* + \mathbf{B}\mathbf{z}^* = \mathbf{c}$, setting $\eta_t = \beta_1 t, \rho_t = \beta_2 t/\lambda_{\max}^{\mathbf{B}}$, we have*

$$R_1(T) \leq G_f^2 \log(T+1)/(2\alpha\beta_1) + \beta_2 D_{\mathbf{z}}^2/2 + \beta_1 D_{\mathbf{x}}^2 \,,$$
$$R^c(T) \leq 2F\lambda_{\max}^{\mathbf{B}} \log(T+1)/\beta_2 + \lambda_{\max}^{\mathbf{B}} D_{\mathbf{z}}^2 + 2\beta_1 \lambda_{\max}^{\mathbf{B}} D_{\mathbf{x}}^2/\beta_2 \,.$$

To guarantee logarithmic regret bounds for both objective and constraints, OADM requires both $f_t$ and $g$ to be strongly convex. FOBOS, COMID, and RDA only require $g$ to be strongly convex although they do not consider linear constraints explicitly.

# 5. Regret Analysis of OADM with $\eta = 0$

We analyze the regret bound when $\eta = 0$. In this case, OADM has the same updates as ADM. For the analysis, we consider $\mathbf{z}_t$ to be the key primal variable, and compute $\hat{\mathbf{x}}_t$ using $\mathbf{z}_t$ so that $\mathbf{A}\hat{\mathbf{x}}_t + \mathbf{B}\mathbf{z}_t = \mathbf{c}$. Since $(\hat{\mathbf{x}}_t, \mathbf{z}_t)$ satisfies the constraints by design, we consider the following regret:

$$R_2(T) = \sum_{t=0}^{T} f_t(\hat{\mathbf{x}}_t) + g(\mathbf{z}_t) - \min_{\mathbf{A}\mathbf{x}+\mathbf{B}\mathbf{z}=\mathbf{c}} \sum_{t=0}^{T} f_t(\mathbf{x}) + g(\mathbf{z}) \,. \quad (21)$$

where $\mathbf{A}\hat{\mathbf{x}}_t + \mathbf{B}\mathbf{z}_t = \mathbf{c}$. A common case we often encounter is when $\mathbf{A} = \mathbf{I}, \mathbf{B} = -\mathbf{I}, \mathbf{c} = \mathbf{0}$, thus $\hat{\mathbf{x}}_t = \mathbf{z}_t$. While $\{\hat{\mathbf{x}}_t, \mathbf{z}_t\}$ satisfies the equality constraint, $(\mathbf{x}_t, \mathbf{z}_t)$ need not satisfy $\mathbf{A}\mathbf{x}_t + \mathbf{B}\mathbf{z}_t - \mathbf{c} = \mathbf{0}$. Thus, in addition to $R_2(T)$, we also consider bounds for $R^c$ as defined in (17).

To guarantee that $\mathbf{A}\hat{\mathbf{x}}_t + \mathbf{B}\mathbf{z}_t = \mathbf{c}$, $\mathbf{A} \in \mathbb{R}^{m\times n_1}$ is feasible, it implicitly requires the assumption $m \leq n_1$. On the other hand, to establish a bound for $R_2$, $\mathbf{A}$ should be full-column rank, i.e., $rank(\mathbf{A}) = n_1$. Therefore, we assume that $\mathbf{A}$ is a square and full rank matrix, i.e., $\mathbf{A}$ is invertible. Let $\lambda_{\min}^{\mathbf{A}}$ be the smallest eigenvalue of $\mathbf{A}\mathbf{A}^T$, then $\lambda_{\min}^{\mathbf{A}} > 0$.

### 5.1. General Convex Functions

The following theorem shows the regret bounds.

**Theorem 5** *Let $\eta = 0$ in OADM and assumptions (1)-(4) and $\mathbf{A}$ is invertible hold. For any $\mathbf{x}^*, \mathbf{z}^*$ satisfying $\mathbf{A}\mathbf{x}^* + \mathbf{B}\mathbf{z}^* = \mathbf{c}$, setting $\rho = \frac{G_f\sqrt{T}}{D_{\mathbf{z}}\sqrt{\lambda_{\min}^{\mathbf{A}}\lambda_{\max}^{\mathbf{B}}}}$, we have*

$$R_2(T) \leq G_f D_{\mathbf{z}}\sqrt{\lambda_{\max}^{\mathbf{B}} T / \lambda_{\min}^{\mathbf{A}}} \,,$$
$$R^c(T) \leq \lambda_{\max}^{\mathbf{B}} D_{\mathbf{z}}^2 + 2F D_{\mathbf{z}}\sqrt{\lambda_{\min}^{\mathbf{A}}\lambda_{\max}^{\mathbf{B}} T}/G_f \;.$$

Without requiring an additional Bregman divergence, $R_2$ achieves the $\sqrt{T}$ bound as $R_1$. While $R_1$ depends on $\mathbf{x}_t$ which may not stay in the feasible set, $R_2$ is defined on $\hat{\mathbf{x}}_t$ which always satisfies the equality constraint. The corresponding algorithm requires finding $\hat{\mathbf{x}}_t$ in each iteration such that $\mathbf{A}\hat{\mathbf{x}}_t = \mathbf{c} - \mathbf{B}\mathbf{z}_t$, which involves solving a linear system. The algorithm will be efficient in some settings, e.g., consensus optimization where $\mathbf{A} = \mathbf{I}$.

### 5.2. Strongly Convex Functions

The following theorem establishes the logarithmic regret bound under the assumption $g$ is $\beta$-strongly convex given in (20).

**Theorem 6** *Let $\eta = 0$ in OADM. Assume that $g(\mathbf{z})$ is $\beta_2$-strongly convex, $\mathbf{A}$ is invertible, and assumptions (1)-(4) hold. Setting $\rho_t = \beta_2 t / \lambda_{\max}^{\mathbf{B}}$, we have*

$$R_2(T) \leq \frac{G_f^2\lambda_{\max}^{\mathbf{B}}}{2\lambda_{\min}^{\mathbf{A}}\beta_2}(\log(T+1)) + \beta_2 D_{\mathbf{z}}^2 \,, \tag{22}$$
$$R^c(T) \leq \lambda_{\max}^{\mathbf{B}} D_{\mathbf{z}}^2 + 2F\lambda_{\max}^{\mathbf{B}}\log(T+1)/\beta_2 \;. \tag{23}$$

Unlike Theorem 4, Theorem 6 shows that OADM can achieve the logarithmic regret bound without requiring $f_t$ to be strongly convex, which is in line with other online learning algorithms for composite objectives.

## 6. Connections to Related Work

In this section, we assume $\eta = 0, \mathbf{A} = \mathbf{I}, \mathbf{B} = -\mathbf{I}, \mathbf{c} = \mathbf{0}$, thus $\mathbf{x} = \mathbf{z}$. The three steps of OADM reduce to

$$\mathbf{x}_{t+1} = \operatorname*{argmin}_{\mathbf{x}} f_t(\mathbf{x}) + \langle \mathbf{y}_t, \mathbf{x} - \mathbf{z}_t\rangle + \frac{\rho}{2}\|\mathbf{x} - \mathbf{z}_t\|^2 \,, \tag{24}$$
$$\mathbf{z}_{t+1} = \operatorname*{argmin}_{\mathbf{z}} g(\mathbf{z}) + \langle \mathbf{y}_t, \mathbf{x}_{t+1} - \mathbf{z}\rangle + \frac{\rho}{2}\|\mathbf{x}_{t+1} - \mathbf{z}\|^2, \tag{25}$$
$$\mathbf{y}_{t+1} = \mathbf{y}_t + \rho(\mathbf{x}_{t+1} - \mathbf{z}_{t+1}) \;. \tag{26}$$

Let $f_t'(\mathbf{x}_{t+1}) \in \partial f_t(\mathbf{x}), g'(\mathbf{z}_{t+1}) \in \partial g(\mathbf{z})$. The first order optimality conditions for (24) and (25) give

$$f_t'(\mathbf{x}_{t+1}) + \mathbf{y}_t + \rho(\mathbf{x}_{t+1} - \mathbf{z}_t) = 0 \,,$$
$$g'(\mathbf{z}_{t+1}) - \mathbf{y}_t - \rho(\mathbf{x}_{t+1} - \mathbf{z}_{t+1}) = 0 \;.$$

Adding them together yields

$$\mathbf{z}_{t+1} = \mathbf{z}_t - \frac{1}{\rho}(f_t'(\mathbf{x}_{t+1}) + g'(\mathbf{z}_{t+1})) \;. \tag{27}$$

OADM can be considered as taking the implicit subgradient of $f_t$ and $g$ at the yet to be determined $\mathbf{x}_{t+1}$ and $\mathbf{z}_{t+1}$. FOBOS has the following update (Duchi & Singer, 2009):

$$\mathbf{z}_{t+1} = \mathbf{z}_t - \frac{1}{\rho}(f_t'(\mathbf{z}_t) + g'(\mathbf{z}_{t+1})) \;.$$

FOBOS takes the explicit subgradient of $f_t$ at current $\mathbf{z}_t$.

As a matter of fact, FOBOS can be considered as an inexact OADM, which linearizes the objective of (24) at $\mathbf{z}_t$ :

$$\mathbf{x}_{t+1} = \operatorname*{argmin}_{\mathbf{x}} \langle f_t'(\mathbf{z}_t) + \mathbf{y}_t, \mathbf{x} - \mathbf{z}_t\rangle + \frac{\tau}{2}\|\mathbf{x} - \mathbf{z}_t\|^2 \;.$$

It has the following closed-form solution:

$$\mathbf{x}_{t+1} = \mathbf{z}_t - \frac{1}{\tau}(f_t'(\mathbf{z}_t) + \mathbf{y}_t) \;. \tag{28}$$

(25) is equivalent to the following scaled form :

$$\mathbf{z}_{t+1} = \operatorname{argmin}_{\mathbf{z}} g(\mathbf{z}) + \frac{\rho}{2}\|\mathbf{x}_{t+1} - \mathbf{z} + \frac{1}{\rho}\mathbf{y}_t\|^2 \;. \tag{29}$$

Let $\rho = \tau$ and $\mathbf{z}_{t+\frac{1}{2}} = \mathbf{x}_{t+1} + \frac{1}{\tau}\mathbf{y}_t$, we get FOBOS (Duchi & Singer, 2009). Furthermore, if $g(\mathbf{z})$ is an indicator function of a convex set $\Omega$, substituting (28) into (29), we have

$$\begin{aligned}\mathbf{z}_{t+1} &= \operatorname{argmin}_{\mathbf{z}\in\Omega} \frac{\rho}{2}\|\mathbf{z}_t - \frac{1}{\tau}f_t'(\mathbf{z}_t) - \mathbf{z}\|^2 \\ &= \mathcal{P}_{\mathbf{z}\in\Omega}\left[\mathbf{z}_t - \frac{1}{\tau}f_t'(\mathbf{z}_t)\right] \;.\end{aligned}$$

We recover the projected gradient descent (Hazan et al., 2007).

## 7. Experimental Results

In this section, we use OADM to solve the generalized lasso problems (Boyd et al., 2010), including lasso (Tibshirani, 1996) and total variation (TV)(Rudin et al., 1992). We present simulation results to show the convergence of objective as well as constraints in OADM. We also compare it with batch ADM and other two online learning algorithms: FOBOS and regularized dual averaging (RDA) in selecting sparse dimension in lasso and recovering data in total variation.

### 7.1. Generalized Lasso

The generalized lasso problem is formulated as follows:

$$\min_{\mathbf{x}} \frac{1}{N}\sum_{t=1}^{N} \|\mathbf{a}_t\mathbf{x} - b_t\|_2^2 + \lambda|\mathbf{D}\mathbf{x}|_1 \,, \tag{30}$$

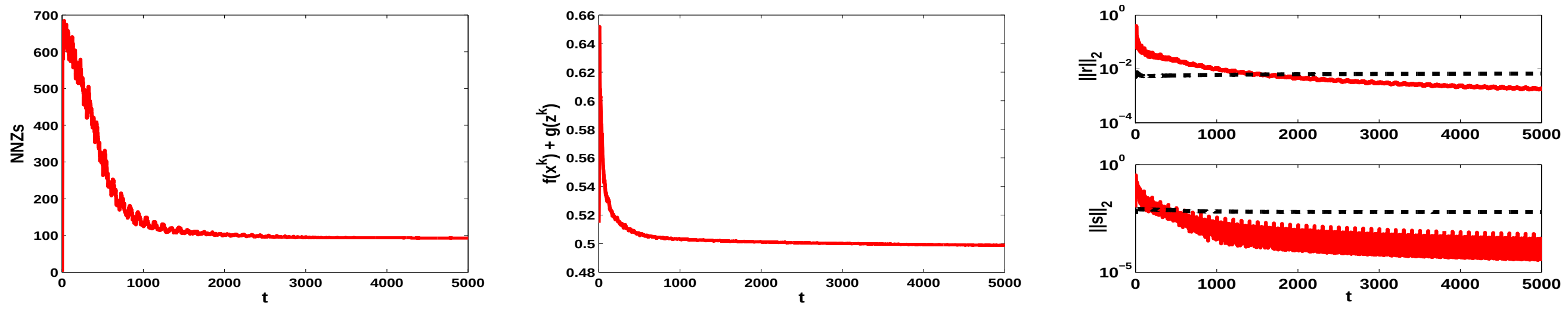

(a) Sparsity. (b) Objective. (c) Constraints (top), primal residual (bottom).

*Figure 1.* The convergence of sparsity, objective value and constraints in OADM with $q = 0.5, \rho = 1, \eta = t$.

where $\mathbf{a}_t \in \mathbb{R}^{1\times n}, \mathbf{x} \in \mathbb{R}^{n\times 1}, \mathbf{D} \in \mathbb{R}^{m\times n}$ and $b_t$ is a scalar. If $\mathbf{D} = \mathbf{I}$, (30) yields the lasso. If $\mathbf{D}$ is an upper bidiagonal matrix with diagonal 1 and off-diagonal $-1$, (30) becomes the total variation. The ADM form of (30) is:

$$\min_{\mathbf{Dx}=\mathbf{z}} \frac{1}{N}\sum_{t=1}^{N} \|\mathbf{a}_t\mathbf{x} - b_t\|_2^2 + \lambda|\mathbf{z}|_1 , \tag{31}$$

where $\mathbf{z} \in \mathbb{R}^{m\times 1}$. The three updates of OADM are:

$$\mathbf{x}_{t+1} = (\mathbf{a}_t^T\mathbf{a}_t + \rho\mathbf{D}^T\mathbf{D} + \eta)^{-1}\mathbf{v} , \tag{32}$$

$$\mathbf{z}_{t+1} = S_{\lambda/\rho}(\mathbf{x} + \mathbf{u}) , \tag{33}$$

$$\mathbf{u}_{t+1} = \mathbf{u}_t + \mathbf{x}_{t+1} - \mathbf{z}_{t+1} , \tag{34}$$

where $\mathbf{u} = \mathbf{y}/\rho$, $\mathbf{v} = \mathbf{a}_t^T b_t + \rho b_t\mathbf{D}^T(\mathbf{z} - \mathbf{u}) + \eta\mathbf{x}$, and $S_{\lambda/\rho}$ denotes the shrinkage operation.

For lasso, the $\mathbf{x}$-update is

$$\mathbf{x}_{t+1} = (\mathbf{v} - (\eta + \rho + \mathbf{a}_t\mathbf{a}_t^T)^{-1}\mathbf{a}_t^T(\mathbf{a}_t\mathbf{v}))/(\eta + \rho) .$$

For total variation, we set $\eta = 0$ so that

$$\mathbf{x}_{t+1} = (\mathbf{Qv} - (\rho + \mathbf{a}_t\mathbf{Q}\mathbf{a}_t^T)^{-1}\mathbf{Q}\mathbf{a}_t^T(\mathbf{a}_t\mathbf{Qv}))/\rho ,$$

where $\mathbf{Q} = (\mathbf{D}^T\mathbf{D})^{-1}$.

In both cases, the three updates (32)-(34) can be done in $O(n)$ flops (Golub & Loan, 1996). In contrast, in batch ADM, the complexity of $x$-update could be as high as $O(n^3)$ or $O(n^2)$ by caching factorizations (Boyd et al., 2010). Here, we do not run them in parallel.

FOBOS and RDA cannot directly solve the TV term. We first reformulate the total variation in the lasso form such that

$$\min_{\mathbf{y}} \frac{1}{N}\sum_{t=1}^{N} \|\mathbf{a}_t\mathbf{D}^{-1}\mathbf{y} - \mathbf{b}\|_2^2 + \lambda|\mathbf{y}|_1 , \tag{35}$$

where $\mathbf{y} = \mathbf{Dx}$. FOBOS and RDA can solve the above lasso problem and get $\mathbf{y}$. $\mathbf{x}$ can be recovered by using $\mathbf{x} = \mathbf{D}^{-1}\mathbf{y}$.

### 7.2. Simulation

Our experiments follow the lasso and total variation examples in Boyd's website,[2] although we modified the codes to accommodate our setup. We first randomly generated $\mathbf{A}$ with $N$ examples of dimensionality $n$. $\mathbf{A}$ is then normalized along the column. Then, a true $\mathbf{x}_0$ is randomly generated with certain sparsity pattern for lasso and TV. $\mathbf{b}$ is calculated by adding gaussian noise to $\mathbf{Ax}_0/N$. In all experiments, $N = 100$, which facilitates the matrix inverse in ADM and will be gone through cyclically in the three online learning algorithms. For lasso, we keep the number of nonzeros (NNZs) $k = 100$ in $\mathbf{x}$ and try different combination of parameters from $n = [1000, 5000]$, $\rho = [0.1, 1, 10]$ and $q = [0.1, 0.5]$ for $\lambda = q \times |\mathbf{A}^T b/N|_\infty$. All experiments are implemented in Matlab.

**Convergence**: We go through the examples 100 times using OADM. Figure 1(a) shows that NNZs converge to some value close to the actual $k = 100$ before $t = 2000$. Figure 1(b) shows the convergence of objective value. In Figure 1(c), the dashed lines are the stopping criteria used in ADM (Boyd et al., 2010). It shows that the equality constraint (top) and primal residual (bottom) are satisfied in the online setting. While the objective converges fast, the equality constraints relatively take more time to be satisfied.

**Sparsity:** We compare NNZs found by batch ADM and three online learning algorithms, including OADM, FOBOS, and RDA. We set $\eta = 1000$ for OADM and $\gamma = 1$ for RDA. For FOBOS, we use a time varying parameter $\rho_t = \rho/\sqrt{t}$. For online learning algorithms, we go through the $N$ examples 100 times. We run the experiment 20 times and the average results are plotted. Due to the limited space, we only show the results for $N = 100, n = 1000, q = 0.5$ in Fig. 2. While ADM and RDA tend to give the sparsest results, OADM seems more conservative and converges to reasonably sparse solutions. Fig.2 shows OADM is closest to the actual NNZs 100. The NNZs in FOBOS is large and oscillates in a big range, which has also been observed in (Xiao, 2010).

[2] http://www.stanford.edu/~boyd/papers/admm/

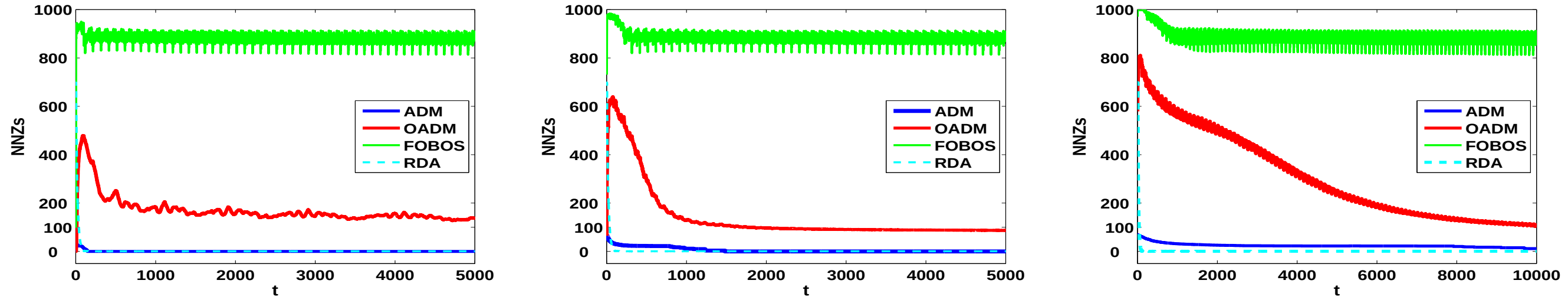

(a) $\rho = 0.1$. (b) $\rho = 1$. (c) $\rho = 10$.

*Figure 2.* The NNZs found by OADM, ADM, FOBOS and RDA with $q = 0.5$. OADM is closest to the actual NNZs.

**Total Variation:** We compare the patterns found by the four algorithms. For all algorithms, $N = 100, n = 1000, \lambda = 0.001$ and $\rho$ is chosen through cross validation. In RDA, $\gamma = 100$. Recall that $\eta = 0$ in OADM. While we use a fixed $\rho$ for OADM and RDA, FOBOS uses $\rho_t = \rho/\sqrt{t}$. Figure 3 shows the three different patterns and results found by the algorithms. ADM seems to follow the pattern with obvious oscillation. OADM is smoother and generally follows the trend of the patterns. For the first two examples, FOBOS works well and the patterns found by RDA tend to be flat. In the last example, both FOBOS and RDA oscillate.

## 8. Conclusions

In this paper, we propose an efficient online learning algorithm named online ADM (OADM). New proof techniques have been developed to analyze the convergence of ADM, which shows that ADM has a $O(1/T)$ convergence rate. Using the proof technique, we establish the regret bounds for the objective and constraint violation for general and strongly convex functions in OADM. Finally, we illustrate the efficacy of OADM in solving lasso and total variation.

## Acknowledgment

The research was supported by NSF CAREER award IIS-0953274, and NSF grants IIS-0916750, IIS-0812183, and IIS-1029711.

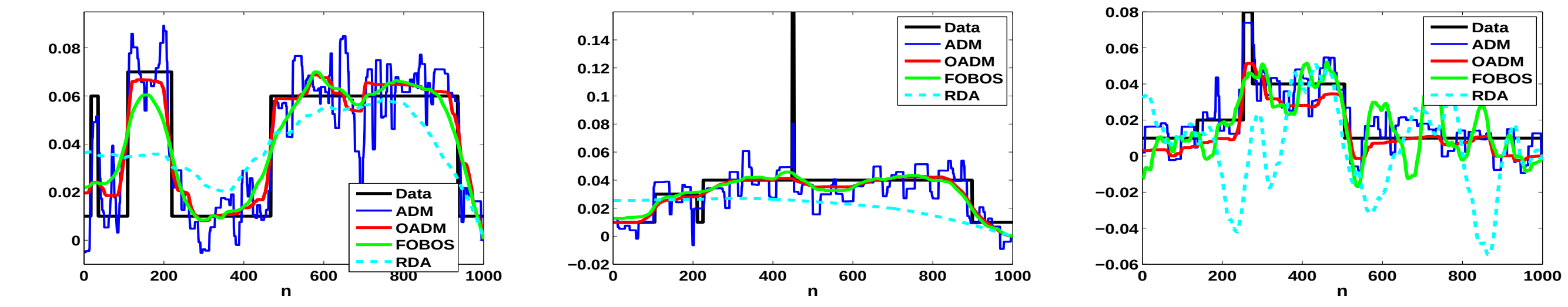

*Figure 3.* The TV patterns found by OADM, ADM, FOBOS and RDA. OADM is the best in recovering the patterns.

## A. Proof of Rate of Convergence of ADM

*Proof:* We start by noting that the VI corresponding to the update of $\mathbf{x}_{t+1}$ in (3) is given by: $\forall \mathbf{x} \in \mathcal{X}$

$$f(\mathbf{x})-f(\mathbf{x}_{t+1})+\langle \mathbf{x}-\mathbf{x}_{t+1}, \mathbf{A}^T\{\mathbf{y}_t+\rho(\mathbf{A}\mathbf{x}_{t+1}+\mathbf{B}\mathbf{z}_t-\mathbf{c})\}\rangle \geq 0 .$$

Using (5), $\forall \mathbf{x} \in \mathcal{X}$

$$\begin{aligned} & f(\mathbf{x}_{t+1}) - f(\mathbf{x}) + \langle \mathbf{x}_{t+1} - \mathbf{x}, \mathbf{A}^T \mathbf{y}_{t+1} \rangle \\ & \leq \rho \langle \mathbf{A}\mathbf{x} - \mathbf{A}\mathbf{x}_{t+1}, \mathbf{B}\mathbf{z}_t - \mathbf{B}\mathbf{z}_{t+1} \rangle , \end{aligned} \tag{36}$$

The VI corresponding to the update of $\mathbf{z}_{t+1}$ in (4) is given by: $\forall \mathbf{z} \in \mathcal{Z}$,

$$g(\mathbf{z})-g(\mathbf{z}_{t+1})+\langle \mathbf{z}-\mathbf{z}_{t+1}, \mathbf{B}^T\{\mathbf{y}_t+\rho(\mathbf{A}\mathbf{x}_{t+1}+\mathbf{B}\mathbf{z}_{t+1}-\mathbf{c})\}\rangle \geq 0 .$$

Using (5), $\forall \mathbf{x} \in \mathcal{X}$

$$g(\mathbf{z}_{t+1}) - g(\mathbf{z}) + \langle \mathbf{z}_{t+1} - \mathbf{z}, \mathbf{B}^T \mathbf{y}_{t+1} \rangle \leq 0 , \tag{37}$$

Adding (36) and (37) and denoting $h(\mathbf{w}) = f(\mathbf{x}) + g(\mathbf{z})$, we have $\forall \mathbf{w} \in \Omega$

$$\begin{aligned} & h(\mathbf{w}_{t+1}) - h(\mathbf{w}) + \langle \mathbf{w}_{t+1} - \mathbf{w}, F(\mathbf{w}_{t+1}) \rangle \\ & \leq \rho \langle \mathbf{A}\mathbf{x} - \mathbf{A}\mathbf{x}_{t+1}, \mathbf{B}\mathbf{z}_t - \mathbf{B}\mathbf{z}_{t+1} \rangle + \frac{1}{\rho} \langle \mathbf{y} - \mathbf{y}_{t+1}, \mathbf{y}_{t+1} - \mathbf{y}_t \rangle . \end{aligned} \tag{38}$$

The first term can be rewritten as

$$\begin{aligned} & 2\langle \mathbf{A}\mathbf{x} - \mathbf{A}\mathbf{x}_{t+1}, \mathbf{B}\mathbf{z}_t - \mathbf{B}\mathbf{z}_{t+1} \rangle \\ & = 2\langle \mathbf{A}\mathbf{x} - \mathbf{c} - (\mathbf{A}\mathbf{x}_{t+1} - \mathbf{c}), \mathbf{B}\mathbf{z}_t - \mathbf{B}\mathbf{z}_{t+1} \rangle \\ & = \|\mathbf{A}\mathbf{x} + \mathbf{B}\mathbf{z}_t - \mathbf{c}\|^2 - \|\mathbf{A}\mathbf{x} + \mathbf{B}\mathbf{z}_{t+1} - \mathbf{c}\|^2 \\ & + \|\mathbf{A}\mathbf{x}_{t+1} + \mathbf{B}\mathbf{z}_{t+1} - \mathbf{c}\|^2 - \|\mathbf{A}\mathbf{x}_{t+1} + \mathbf{B}\mathbf{z}_t - \mathbf{c}\|^2 . \end{aligned} \tag{39}$$

The second term in (38) is equivalent to

$$\begin{aligned} & 2\langle \mathbf{y} - \mathbf{y}_{t+1}, \mathbf{y}_{t+1} - \mathbf{y}_t \rangle \\ & = \|\mathbf{y} - \mathbf{y}_t\|^2 - \|\mathbf{y} - \mathbf{y}_{t+1}\|^2 - \|\mathbf{y}_t - \mathbf{y}_{t+1}\|^2 . \end{aligned} \tag{40}$$

Substituting (39) and (40) into (38) and summing over $t$,

$$\sum_{t=1}^{T} [h(\mathbf{w}_t) - h(\mathbf{w}) + \langle \mathbf{w}_t - \mathbf{w}, F(\mathbf{w}_t) \rangle] \leq L , \tag{41}$$

where the constant $L = \frac{\rho}{2}\|\mathbf{A}\mathbf{x} - \mathbf{c}\|_2^2 + \frac{1}{2\rho}\|\mathbf{y}\|^2$. Recall that $h(\tilde{w})$ is a convex function of $\tilde{w}$. Further, from the definition of $F(\tilde{\mathbf{w}})$, $\langle \tilde{\mathbf{w}} - \mathbf{w}, F(\tilde{\mathbf{w}}) \rangle$ is a convex function of $\tilde{\mathbf{w}}$. Dividing both sides of (41) by $T$, recalling that $\bar{\mathbf{w}}_T = \frac{1}{T}\sum_{t=1}^{T} \mathbf{w}_t$, and using Jensen's inequality, we have

$$\begin{aligned} & h(\bar{\mathbf{w}}_T) - h(\mathbf{w}) + \langle \bar{\mathbf{w}}_T - \mathbf{w}, F(\bar{\mathbf{w}}_T) \rangle \\ & \leq \frac{1}{T}\sum_{t=1}^{T} h(\mathbf{w}_t) - h(\mathbf{w}) + \frac{1}{T}\sum_{t=1}^{T} \langle \mathbf{w}_t - \mathbf{w}, F(\mathbf{w}_t) \rangle \\ & \leq \frac{L}{T} = O\left(\frac{1}{T}\right) , \end{aligned}$$

which establishes convergence rate for ADM. ■